\pdfoutput=1

\documentclass[11pt]{article}

\usepackage[final]{coling}

\usepackage{times}
\usepackage{latexsym}

\usepackage[T1]{fontenc}

\usepackage[utf8]{inputenc}

\usepackage{microtype}

\usepackage{inconsolata}

\usepackage{graphicx}

\usepackage{expex}
\lingset{
    aboveglftskip=1pt,
    interpartskip=\baselineskip,
    textoffset=9pt
}

\usepackage{devanagari}
\usepackage{epsfig}

\DeclareTextFontCommand{\suitename}{\ttfamily\hyphenchar\font=45\relax}

\usepackage{amsmath}
\newcommand{\tildemacron}[1]{\stackrel{\sim}{\smash{\bar{#1}}\vphantom{#1}}}

\usepackage{float}

\title{Controlled Evaluation of Syntactic Knowledge in Multilingual Language Models}

\author{Daria Kryvosheieva \\
  MIT \\
  \texttt{daria\_k@mit.edu} \\\And
  Roger Levy \\
  MIT \\
  \texttt{rplevy@mit.edu} \\}

\begin{document}
\maketitle
\begin{abstract}
Language models (LMs) are capable of acquiring elements of human-like syntactic knowledge. Targeted syntactic evaluation tests have been employed to measure how well they form generalizations about syntactic phenomena in high-resource languages such as English. However, we still lack a thorough understanding of LMs' capacity for syntactic generalizations in low-resource languages, which are responsible for much of the diversity of syntactic patterns worldwide. In this study, we develop targeted syntactic evaluation tests for three low-resource languages (Basque, Hindi, and Swahili) and use them to evaluate five families of open-access multilingual Transformer LMs. We find that some syntactic tasks prove relatively easy for LMs while others (agreement in sentences containing indirect objects in Basque, agreement across a prepositional phrase in Swahili) are challenging. We additionally uncover issues with publicly available Transformers, including a bias toward the habitual aspect in Hindi in multilingual BERT and underperformance compared to similar-sized models in XGLM\textsubscript{4.5B}.
\vspace{3pt}

\includegraphics[width=1.25em,height=1.25em]{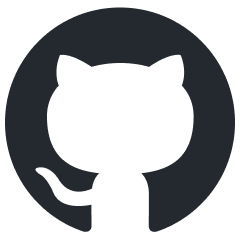}\hspace{.75em}\parbox{\dimexpr\linewidth-2\fboxsep-2\fboxrule}{\href{https://github.com/dariakryvosheieva/syntactic_generalization_multilingual}{\ttfamily dariakryvosheieva/syntactic\_\linebreak generalization\_multilingual}}
\vspace{3pt}
\end{abstract}

\section{Introduction}
\label{sec:introduction}

There is a substantial body of work dedicated to evaluating the linguistic knowledge of language models. Popular evaluation methodologies include:
\begin{itemize}
\item probing, i.e., predicting linguistic properties from a network's internal activations \cite{giulianelli-2018};
\item classifying sentences as grammatically acceptable or unacceptable \cite{warstadt-2019};
\item targeted syntactic evaluation (TSE), a method based on comparing LM-assigned probabilities of minimally different sequences \cite{marvin-2018}.
\end{itemize}

To date, most of the research investigating the linguistic knowledge of LMs has concentrated on high-resource languages such as English \citep{lin-2019, warstadt-2020, hu-2020}, German \citep{mueller-2020, zaczynska-2020}, Spanish \citep{perezmayos-2021, bel-2024}, Italian \citep{trotta-2021, miaschi-2022}, Chinese \citep{wang-2021, xiang-2021, zheng-2023}, and Japanese \citep{futrell-2018, someya-2023, someya-2024}. However, efforts have also been made to include less prominent languages. Notably, \citet{torrobahennigen-2020} and \citet{stanczak-2022} probed masked LMs for morphosyntactic attributes of words across 36 and 43 languages, respectively. Acceptability benchmarks have been developed for North Germanic languages by \citet{volodina-2021}, 
\citet{jentoft-2023},
\citet{nielsen-2023}, and \citet{zhang-2024}. TSE has been applied to Hebrew \cite{gulordava-2018}, Norwegian \cite{kobzeva-2023}, and Indonesian and Tamil \cite{leong-2023}.

We believe that evaluating LMs' linguistic knowledge across a diverse range of languages is crucial for developing a comprehensive picture of how they form linguistic generalizations. Assessing `off-the-shelf' LMs in lower-resourced languages offers a further benefit of diagnosing limitations and challenges these models may face due to issues like insufficient training data, model biases, or difficulties in capturing particular linguistic features. We recognize TSE's advantage of focusing on one combination of linguistic phenomenon and sentence structure at a time, which enables a fine-grained analysis of how performance depends on the structure and complexity of input sentences. While many prior TSE studies considered LMs based on RNN and LSTM architectures \citep{linzen-2016, wilcox-2021}, we focus on modern LMs based on the Transformer architecture, which is the current state of the art. Therefore, we conduct three TSE case studies benchmarking publicly available Transformer LMs on distinctive morphosyntactic phenomena in low-resource languages—auxiliary verb agreement in Basque, split ergativity in Hindi, and noun class agreement in Swahili.

We find that LMs mostly do well on agreement in Basque, with errors linked to the presence of an indirect object in a sentence, and almost always succeed in selecting the correct aspectual form of the verb based on the presence or absence of the ergative clitic in Hindi, with the exception of multilingual BERT, which prefers the habitual aspect regardless of its grammaticality. However, LMs struggle to agree predicates with the noun class of their subjects in Swahili. Performance on our tests has a positive relationship with model size, but XGLM\textsubscript{4.5B} systematically underperforms similar-sized models for reasons possibly including the lack of low-resource upsampling and the `curse of multilinguality'. Syntactically complex attractor phrases weaken performance in Swahili but not in Hindi.

\section{Evaluation Paradigm}
\label{sec:evaluation-paradigm}

Following existing work on TSE, we organize our evaluation materials in \textit{minimal pairs}—pairs of minimally different sentences such that one is grammatically acceptable and the other one is ungrammatical because it violates a specific linguistic rule. Below we provide an English example where sentences (A) and (B) differ by one property—the plurality of the copula. In sentence (B), the plural copula `are' does not agree with the singular subject `The teacher', rendering the sentence ungrammatical.
\ex[exno=A, glhangstyle=none]
The teacher \underline{is} good.
\xe
\ex~[exno=B, glhangstyle=none]
\ljudge*
The teacher \underline{are} good.
\xe
In all our minimal pairs, the sentences differ by one word whose grammaticality can be determined from the left context. We call the left context the \textit{condition} and the rest of the sentence the \textit{target}: in the example above, `The teacher' is the condition, `is good.' is the grammatical target, and `are good.' is the ungrammatical target. We use the \texttt{minicons} Python library \cite{misra-2022} to compute the conditional log-probabilities of the grammatical and ungrammatical targets given the condition, expecting a model to assign a more positive log-probability to the grammatical target if it has learned the rule correctly. We group minimal pairs into \textit{test suites}, each of which assesses models' knowledge of one phenomenon in sentences of a unified structure, and report a model's accuracy on a test suite as the proportion of minimal pairs for which it assigns higher log-probability to the grammatical target.

\section{Test Suites}
\label{sec:test-suites}

We consider three low-resource languages from different language families: Basque (isolate), Hindi (Indo-European), and Swahili (Niger-Congo). In each of them, we focus on one characteristic morphosyntactic phenomenon: in Basque, auxiliary verb agreement (\textsection \ref{sec:auxiliary-verb-agreement-in-basque}); in Hindi, split ergativity (\textsection \ref{sec:split-ergativity-in-hindi}); in Swahili, noun class agreement (\textsection \ref{sec:noun-class-agreement-in-swahili}). We generate synthetic test suites (\textsection \ref{sec:data-generation}) and perform human validation to verify that the generated minimal pairs represent a genuine contrast in grammatical acceptability (\textsection \ref{sec:human-validation}).

\subsection{Auxiliary verb agreement in Basque}
\label{sec:auxiliary-verb-agreement-in-basque}

The Basque verbal agreement system is more complex than that of most other languages because Basque verbs must agree with all of their arguments—not just the subject but also the direct and indirect object if present in the sentence. Verbs typically consist of a non-finite stem and an auxiliary that carries agreement morphology. For each possible set of arguments—Subject (S); Subject and Direct Object (S DO); Subject, Indirect Object, and Direct Object (S IO DO); Indirect Object and Subject (IO S)—we separately test the agreement of the auxiliary with each argument, resulting in a total of eight test suites. Example 1 below presents a minimal pair from the \suitename{basque-DO-S\_DO\_V\_AUX} test suite, which tests the agreement of the auxiliary with the direct object in sentences of the form `S DO V AUX':

\vbox{
\footnotesize
\ex[exno=1.a, glhangstyle=none]
\begingl
    \gla Saltzaileak tomateak prestatu \underline{zituen}. //
    \glb salesman.\textsc{erg}.\textsc{sg} tomato.\textsc{abs}.\textsc{pl} prepare.\textsc{pst.pfv} \textsc{pst.3sg>3pl} //
    \glft `The salesman prepared the tomatoes.' //
\endgl
\xe
\ex~[exno=1.b, glhangstyle=none]
\ljudge*
\begingl
    \gla Saltzaileak tomateak prestatu \underline{zuen}. //
    \glb salesman.\textsc{erg}.\textsc{sg} tomato.\textsc{abs}.\textsc{pl} prepare.\textsc{pst.pfv} \textsc{pst.3sg>3sg} //
    \glft (ungrammatical) //
\endgl
\xe
}

\vspace{11pt}

\noindent In sentence (1.b), the auxiliary \textit{zuen} correctly agrees with the singular subject. However, its direct object specification (singular) mismatches the actual number of the direct object (plural). We note that if the subject and the direct object in this example shared the same number, a model's preference for the correct auxiliary \textit{zituen} would be compatible with an incorrect heuristic: associating the infix \textit{-it-} with a plural subject rather than a plural direct object. To control for this, we generate minimal pairs ensuring that the number of the focused argument (here, the direct object) differs from the number of the other arguments.

\subsection{Split ergativity in Hindi}
\label{sec:split-ergativity-in-hindi}

Hindi exhibits ergative-absolutive alignment in the perfective aspect and nominative-accusative alignment otherwise. The subject receives the ergative clitic \textit{{\dn n\?}} (\textit{ne}) if and only if the sentence is perfective and transitive. Thus, given an input of the form `S \textit{{\dn n\?}} O' (`Subject \textit{ne} Object'), models should prefer a perfective verb form over a non-perfective form (habitual or progressive). Conversely, given `S O' without \textit{{\dn n\?}}, non-perfective forms should be preferred.

We experiment with varying complexities of the direct object noun phrase, which stands between \textit{{\dn n\?}} and the verb. The direct object structures include:
\begin{enumerate}
\item Noun (e.g., `carrot')
\vspace{-7pt}
\item Possessive pronoun + noun (`their carrot')
\vspace{-7pt}
\item Possessive pronoun + noun\textsubscript{1} + genitive marker + noun\textsubscript{2} (`their friend's carrot')
\end{enumerate}
For each of these direct object structures, we prepare a test suite with \textit{{\dn n\?}} (ergative-absolutive), where we expect models to prefer a perfective verb, and one without \textit{{\dn n\?}} (nominative-accusative), where we expect them to prefer a non-perfective verb. We select the habitual aspect as the alternative to the perfective in our minimal pairs to avoid possible unnatural combinations of the progressive aspect with stative verbs. Example 2 shows a minimal pair from the \suitename{hindi-S\_PossPRN\_O\_V} test suite, which tests whether models prefer the habitual aspect over the perfective aspect in sentences with a `possessive pronoun + noun' direct object that do not include the \textit{{\dn n\?}} marker.

\vbox{
\footnotesize
\ex[exno=2.a, glhangstyle=none]
\begingl
    \gla {\dn sA\1w} {\dn inkF} {\dn gAjr} {\dn {\underline{KAtA} h\4}.} //
    \glb s$\tildemacron{\text{a}}$ṛ inkī gājar {khātā hai} //
    \glc bull\textsc{.m.sg}  their\textsc{.f.sg} carrot\textsc{.f.sg} eat\textsc{.hab.prs.m.sg} //
    \glft `The bull eats their carrot.' //
\endgl
\xe
\ex~[exno=2.b, glhangstyle=none]
\ljudge*
\begingl
    \gla {\dn sA\1w} {\dn inkF} {\dn gAjr} {\dn {\underline{KAyA} h\4}.} //
    \glb s$\tildemacron{\text{a}}$ṛ inkī gājar {khāyā hai} //
    \glc bull\textsc{.m.sg} their\textsc{.f.sg} carrot\textsc{.f.sg} eat\textsc{.pfv.prs.m.sg} //
    \glft (ungrammatical) //
\endgl
\xe
}

\subsection{Noun class agreement in Swahili}
\label{sec:noun-class-agreement-in-swahili}

Swahili has a two-dimensional noun class system based on semantic meaning and number, comprising 18 classes in total. Every noun carries a prefix corresponding to its class, although in some cases the prefix may be zero. Typically, a verb must agree with the noun class of its subject, an adjective must agree with the class of the noun it modifies, and the preposition equivalent to English `of' in possessive constructions (`X of Y') must agree with the class of the possessee (X).

We test the agreement of verbal and adjectival predicates with their subjects in sentences where the subject is modified by a possessor prepositional phrase, which stands between the subject and the predicate and thus serves as a potential distractor. We vary the complexity of the possessor:
\begin{enumerate}
\item Noun (e.g., `scientists')
\vspace{-7pt}
\item Noun + demonstrative (`these scientists')
\vspace{-7pt}
\item Noun + demonstrative + adjective (`these old scientists')
\vspace{-7pt}
\item Noun + demonstrative + adjective + relative verb (`these old scientists that jumped')
\end{enumerate}
We independently vary whether the predicate is a verb or an adjective. To rule out cases where the selection of the correct prefix results from attending to the wrong noun, we ensure that the possessor's noun class is different from that of the subject. Example 3 below is taken from the \suitename{swahili-N\_of\_Poss\_D\_ni\_A} test suite, which tests the agreement of an adjectival predicate with a subject modified by a `noun + demonstrative' possessor.

\vbox{
\footnotesize
\ex[exno=3.a, glhangstyle=none]
\begingl
    \gla Nyumba za wanasayansi hawa wazee ni nyekundu. //
    \glb ny-umba z-a w-anasayansi hawa wa-zee ni ny-ekundu //
    \glc 10-house 10-of 2-scientist 2.this 2-old is 10-red //
    \glft `The houses of these old scientists are red.' //
\endgl
\xe
\ex~[exno=3.b, glhangstyle=none]
\ljudge*
\begingl
    \gla Nyumba za wanasayansi hawa wazee ni wekundu. //
    \glb ny-umba z-a w-anasayansi hawa wa-zee ni w-ekundu //
    \glc 10-house 10-of 2-scientist 2.this 2-old is 2-red //
    \glft (ungrammatical) //
\endgl
\xe
}

\noindent Here, the predicate adjective `red' (\textit{-ekundu}) can only refer to the subject `houses' (\textit{nuymba}), not the possessor `scientists' (\textit{wanasayansi}). Therefore, the correct noun class prefix for the adjective is the one corresponding to `houses' (class 10), not `scientists' (class 2).

\subsection{Data generation}
\label{sec:data-generation}

We adopt the approach from the BLiMP paper \cite{warstadt-2020} to generate artificial sentences. For each language, we manually assemble a vocabulary of approximately 300 words, annotating them with relevant syntactic, morphological, and semantic properties. We then prepare generation scripts that randomly sample words from the vocabulary according to predefined templates specifying sentence structures and required word properties. Grouping words by semantic categories allows us to generate sentences that sound more or less plausible: for instance, we avoid sampling inanimate nouns as subjects of active verbs, or inedible nouns as objects of the verb 'to eat'. Inflections that follow highly regular patterns—like Basque case endings—are added to stems via rule-based algorithms. Inflected forms that are less regular—such as Hindi case forms—are listed as special entries in the vocabulary. Using this procedure, we generate 1,000 minimal pairs per test suite.

\subsection{Human validation}
\label{sec:human-validation}

We designed a human experiment on Prolific that mirrored the task given to LMs. For each language, we presented self-reported L1 speakers with a subset of minimal pairs we generated in that language and asked them to select the more grammatically acceptable sentence in every pair. The datasets presented to speakers consisted of five randomly sampled minimal pairs from every test suite associated with the language, resulting in a total of 40 pairs for Basque and Swahili and 30 pairs for Hindi. To ensure that the participants were legitimate speakers of the languages, we additionally included two control minimal pairs in each dataset. In the control pairs, the grammatical sentence was taken from a published text (Basque: \textit{Euskaltzaindiaren Hiztegia} [Dictionary of the Royal Academy of the Basque Language]; Hindi: \textit{Basic Hindi} by Rajiv Ranjan; Swahili: BBC Swahili news reports), and the ungrammatical sentence was created by altering the target word to a nonsensical form (Basque: rewriting the auxiliary backwards; Hindi: replacing the suffix on the aspectual participle with a nonsensical syllable; Swahili: replacing the class prefix on the verb with a nonsensical syllable). The order of the sentences was shuffled. All participants were paid for 20 minutes of work at a rate of \$15.50/hour, but only submissions that selected the grammatical sentence in both control trials were considered for analysis. We stopped recruiting new participants once we received ten such submissions per language.

We used BLiMP's threshold for the inclusion of a test suite in the LM experiment: in at least four out of five minimal pairs, the majority of reviewers must have selected the intended-grammatical sentence. All test suites except two Swahili test suites passed the threshold. For the test suites that passed the threshold, we report human accuracy scores (the percentage of times that validators selected the grammatical sentence) together with LM accuracy scores in Figure \ref{fig:results}.

\section{Models}
\label{sec:models}

We evaluated five open-access multilingual Transformers from Hugging Face: three autoregressive models—mGPT, BLOOM, and XGLM—and two masked models—multilingual BERT (mBERT) and XLM-RoBERTa (XLM-R). Each of these models except mBERT is available in several versions differing by size; we evaluated all versions. For an overview of the models and their versions, see Appendix \ref{sec:appendix}.

\section{Results}
\label{sec:results}

\begin{figure*}[h!]
\centering
\includegraphics[width=1\linewidth]{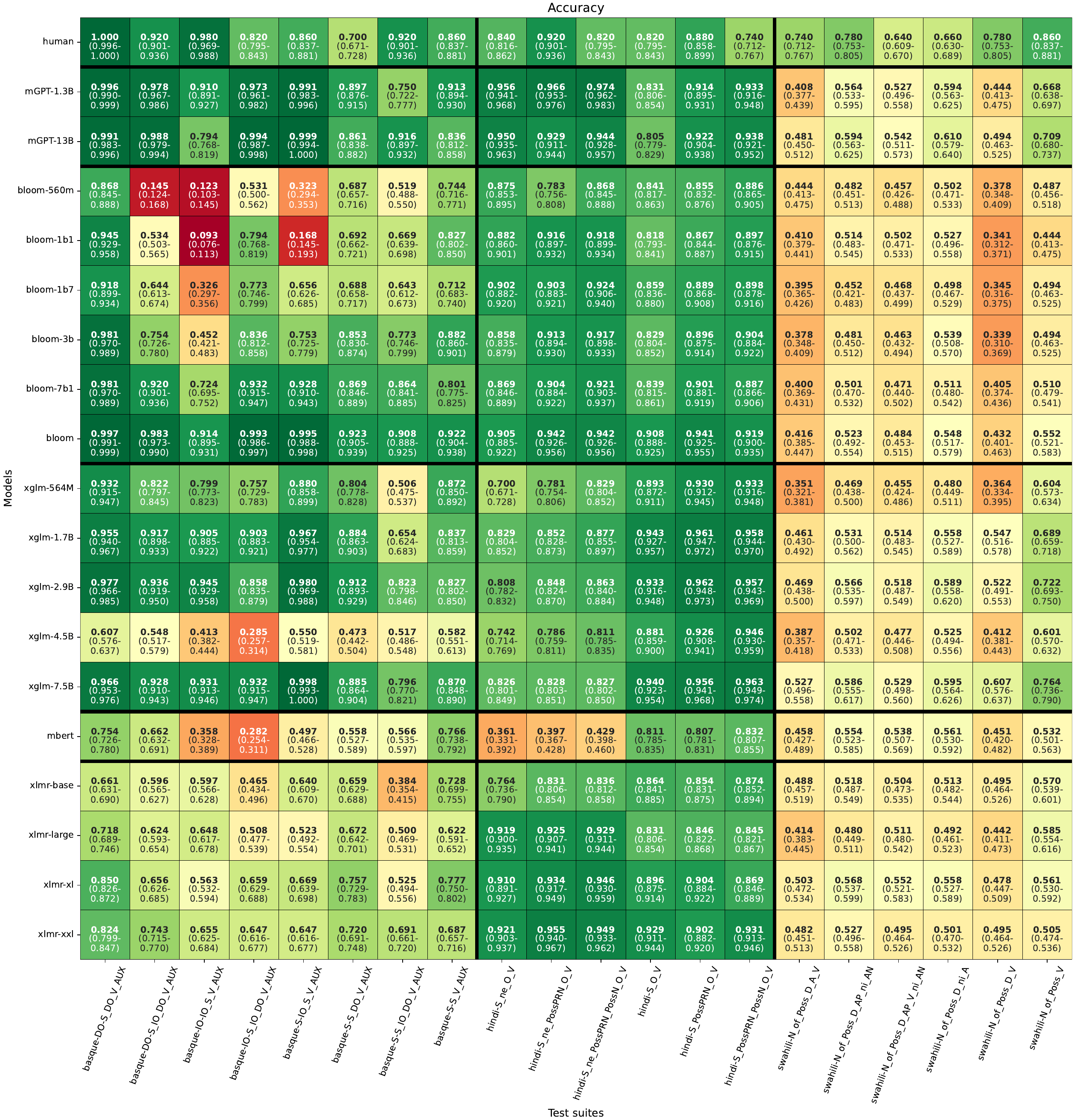}
\caption{Accuracy scores of the models (vertical axis) on our test suites (horizontal axis). In each cell, the bolded value denotes the fraction of minimal pairs in which the model selected the grammatical target, while values in parentheses denote the left and right 95\% confidence intervals. The expectation for random guessing is 0.5.}
\label{fig:results}
\end{figure*}

We present evaluation results in Figure \ref{fig:results} and provide an overview by language in \textsection \ref{sec:overview-by-language}. For models that come in several versions, we analyze the relationship between performance on our test suites and the number of parameters (\textsection \ref{sec:number-of-parameters-and-performance}). For test suite classes where we varied the complexity of intervening phrases (split ergativity in Hindi and noun class agreement in Swahili), we analyze the relationship between performance and complexity of the intervening constituent (\textsection \ref{sec:robustness-to-intervening-content}).

\subsection{Overview by language}
\label{sec:overview-by-language}

We find that LMs perform the best on split ergativity in Hindi (average accuracy score 0.873), followed by auxiliary verb agreement in Basque (0.741) and noun class agreement in Swahili (0.504). The top-performing model is mGPT\textsubscript{13B} overall (average accuracy 0.815), mGPT\textsubscript{1.3B} on Basque (0.926), XLM-R\textsubscript{XXL} on Hindi (0.931), and XGLM\textsubscript{7.5B} on Swahili (0.601).

\paragraph{Basque} Multiple models, namely mGPT\textsubscript{1.3B}, mGPT\textsubscript{13B}, BLOOM\textsubscript{7.1B}, BLOOM\textsubscript{176B}, XGLM\textsubscript{1.7B}, XGLM\textsubscript{2.9B}, XGLM\textsubscript{7.5B}, and XLM-R\textsubscript{XXL}, perform significantly above chance ($p < 0.05$ using a one-sided binomial test) on all Basque test suites. At the same time, at least one model performs significantly below chance ($p < 0.05$) on the \texttt{basque-S-S\_DO\_V\_AUX} test suite as well as all test suites containing indirect objects (including IO agreement and agreement with other arguments in the presence of an IO). Our observation that IOs confuse Transformer LMs is consistent with the observation of \citet{ravfogel-2018} that an LSTM-based classifier trained on a morphologically annotated Wikipedia corpus struggles to predict the number of dative arguments of Basque verbs. Ravfogel et al. hypothesized that the low recall scores they obtained on the dative argument plurality prediction task were caused by the relative rarity of dative nouns in the corpus their LSTM was trained on. Our examination of the Universal Dependencies \cite{nivre-2020} treebank for Basque supports the hypothesis that dative nouns (IOs) are relatively infrequent in the Basque language: the treebank contains a total of 8,595 subject noun phrases, 7,508 direct objects, and 1,021 indirect objects, which means that indirect objects are approximately eight times less frequent than subjects and seven times less frequent than direct objects. We thus hold it plausible that the low frequency of IOs indeed hinders neural networks' ability to learn how they fit into sentences. We note that sentences containing IOs use completely different forms of auxiliaries from sentences without IOs because Basque auxiliaries follow different conjugation paradigms for each set of arguments. For this reason, LMs cannot use information from the more frequent sentences without IOs to infer the conjugations of auxiliaries used in sentences with IOs. Having to learn those conjugations from scratch in a low-frequency setting is what we presume reduces performance.

\paragraph{Hindi} Nearly all models perform significantly above chance on all Hindi test suites. The only exception is mBERT, which performs significantly below chance on the three ergative-absolutive test suites. Multilingual BERT displays a bias toward the habitual aspect, preferring it even in constructions where it is ungrammatical because the perfective aspect is expected. If the Universal Dependencies Hindi PUD treebank is representative of broader usage dynamics of aspectual forms in Hindi, this bias is not explained by the relative frequencies of perfective and habitual verbs: in the treebank, perfective forms are slightly more frequent than habitual forms both overall (641 vs. 515) and specifically in `subject-object-verb' clauses (284 vs. 190).

\paragraph{Swahili} LM scores on Swahili test suites concentrate within 0.2 from the random guessing baseline of 0.5, with the exception of three scores above 0.7, obtained by mGPT\textsubscript{13B}, XGLM\textsubscript{2.9B}, and XGLM\textsubscript{7.5B} on the \texttt{swahili-N\_of\_Poss\_V} test suite. Only one model (XGLM\textsubscript{7.5B}) performs significantly above chance on every Swahili test suite, while three models (BLOOM\textsubscript{560M}, BLOOM\textsubscript{1.7B}, BLOOM\textsubscript{7.1B}) never perform significantly above chance on Swahili test suites. The simplest agreement task we consider, where the subject and a verbal predicate are separated by a simple `preposition + noun' prepositional phrase, proves to be the easiest for LMs: 10 out of 18 LMs achieve their highest Swahili score on this test suite, this is the only Swahili test suite on which some LM scores exceed 0.7, and it also has the highest number of models performing significantly above chance among Swahili test suites. However, performance plummets once demonstratives and adjectives are added to the intervening phrase (see \textsection \ref{sec:robustness-to-intervening-content} for details).

\subsection{Number of parameters and performance}
\label{sec:number-of-parameters-and-performance}

For Transformer LMs, capability is known to improve with size. For example, \citet{kaplan-2020} found a power-law relationship between number of parameters and crossentropy loss on the test set, and \citet{tay-2023} found a positive linear relationship between number of parameters and an aggregate of GLUE, SuperGLUE, and SQuAD scores. At the same time, \citet{warstadt-2020} argued that GPT-2 model size has no significant effect on BLiMP accuracy.

\begin{figure}[h!]
\centering
\includegraphics[width=1\linewidth]{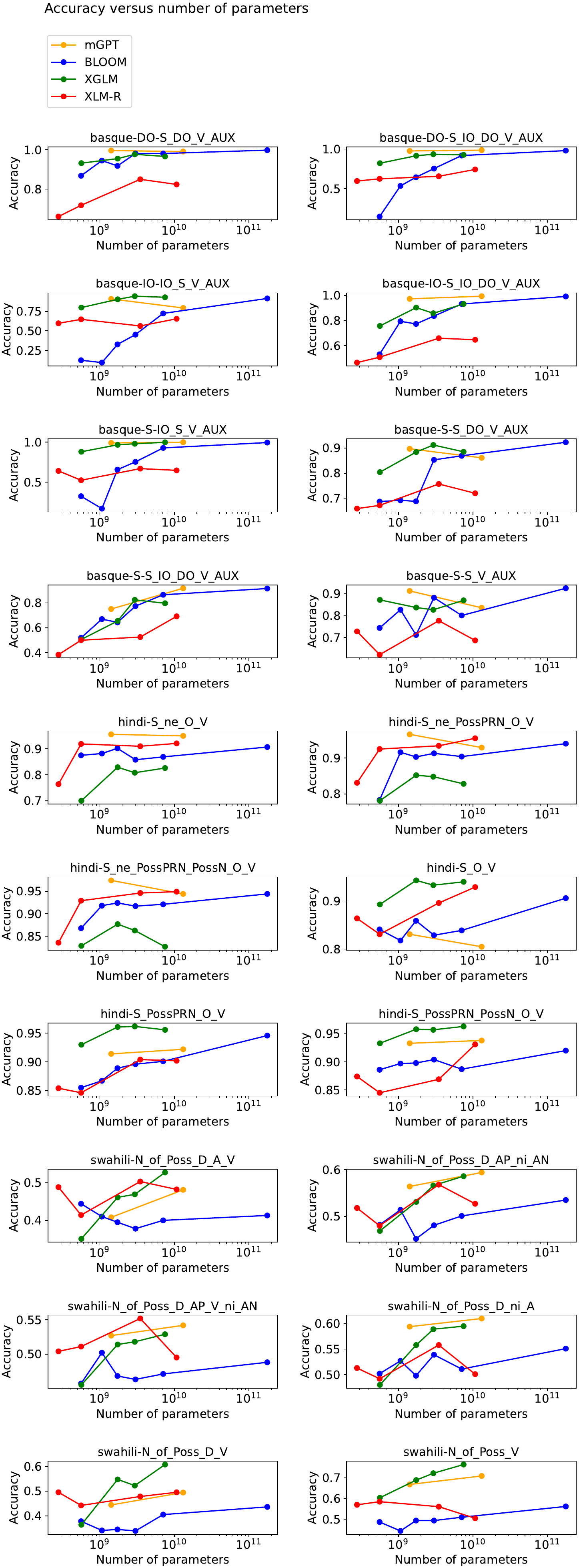}
\caption{Accuracy as a function of parameter count for each model family and test suite.}
\label{fig:performance-vs-size}
\end{figure}

The fact that mGPT, BLOOM, XGLM, and XLM-R are each available in multiple versions differing by size but having the same architecture and trained on the same corpus provides us with ground for a controlled analysis of the relationship between model size and performance on our test suites. We plot accuracy as a function of parameter count for each test suite and model family and present the plots in Figure \ref{fig:performance-vs-size}. In the XGLM family, we exclude XGLM\textsubscript{4.5B} because it was trained on a different variant of the corpus covering more languages. We use linear regression to obtain slopes of best fit lines, which we report in Table \ref{tab:4}.\footnote{For mGPT, the best fit line is simply the line connecting the accuracy scores of mGPT\textsubscript{1.3B} and mGPT\textsubscript{13B}.} We observe that the majority of slopes are positive, including 12 out of 20 slopes in the case of mGPT, all slopes in the case of BLOOM, all slopes except the one corresponding to the \suitename{hindi-S\_ne\_PossPRN\_PossN\_O\_V} test suite in the case of XGLM, and all slopes except three corresponding to Swahili test suites in the case of XLM-R. Furthermore, we find that the average slope over the four model families is positive for all test suites except \suitename{basque-S-S\_V\_AUX}. Contrary to the finding in the BLiMP paper, this suggests a positive relationship between model size and accuracy.

XGLM\textsubscript{4.5B} shows the poorest performance among XGLM versions on 10 out of 18 test suites (including 7 out of 8 Basque test suites) and the second-poorest performance after the smallest version (XGLM\textsubscript{564M}) on the remaining 8 test suites. Additionally, it is outperformed on most test suites by similar-sized non-XGLM models (mGPT\textsubscript{1.3B}, BLOOM\textsubscript{1.7B}, BLOOM\textsubscript{3B}, BLOOM\textsubscript{7.1B}, XLM-R\textsubscript{XL}). Available information about the model's training procedure and dataset is limited, apart from the fact that it was trained on all 134 languages featured in the CC100 XL corpus \cite{lin-2022}, by contrast to other XGLM models, which were trained on a 30-language subset sampled from the same corpus with an upsampling of lower-resourced languages. Therefore, the reasons behind the model's underperformance remain unclear, but we conjecture that the underperformance could be attributed to the lack of low-resource upsampling—in particular, this would explain the low performance on Basque, since the CC100 XL corpus contains much less Basque data (0.35 GiB) than the corpora used to train BLOOM (2.2 GiB) and XLM-R (2.0 GiB)—and the `curse of multilinguality' \cite{conneau-2020}, the phenomenon that training a small model on many languages leads to performance degradation if the number of languages exceeds a certain threshold. We note that XGLM\textsubscript{4.5B} supports the largest number of languages among all models we consider.

\begin{table*}[h!]
\footnotesize
\centering
\begin{tabular}{c|ccccc}
\textbf{Test suite} & \textbf{mGPT} & \textbf{BLOOM} & \textbf{XGLM} & \textbf{XLM-R} & \textbf{Average} \\
\hline
\suitename{basque-DO-S\_DO\_V\_AUX} & -0.043 & 0.035 & 0.371 & 1.270 & 0.408 \\
\suitename{basque-DO-S\_IO\_DO\_V\_AUX} & 0.086 & 0.231 & 1.057 & 1.294 & 0.667 \\
\suitename{basque-IO-IO\_S\_V\_AUX} & -0.992 & 0.339 & 1.377 & 0.353 & 0.269 \\
\suitename{basque-IO-S\_IO\_DO\_V\_AUX} & 0.180 & 0.131 & 1.878 & 1.518 & 0.927 \\
\suitename{basque-S-IO\_S\_V\_AUX} & 0.068 & 0.258 & 1.298 & 0.586 & 0.553 \\
\suitename{basque-S-S\_DO\_V\_AUX} & -0.308 & 0.098 & 0.750 & 0.502 & 0.261 \\
\suitename{basque-S-S\_IO\_DO\_V\_AUX} & 1.420 & 0.128 & 3.475 & 2.437 & 1.865 \\
\suitename{basque-S-S\_V\_AUX} & -0.659 & 0.075 & 0.192 & 0.057 & -0.084 \\
\suitename{hindi-S\_ne\_O\_V} & -0.051 & 0.016 & 1.196 & 0.785 & 0.486 \\
\suitename{hindi-S\_ne\_PossPRN\_O\_V} & -0.316 & 0.034 & 0.302 & 0.753 & 0.193 \\
\suitename{hindi-S\_ne\_PossPRN\_PossN\_O\_V} & -0.257 & 0.019 & -0.313 & 0.647 & 0.024 \\
\suitename{hindi-S\_O\_V} & -0.222 & 0.041 & 0.436 & 0.775 & 0.258 \\
\suitename{hindi-S\_PossPRN\_O\_V} & 0.068 & 0.035 & 0.215 & 0.488 & 0.202 \\
\suitename{hindi-S\_PossPRN\_PossN\_O\_V} & 0.043 & 0.014 & 0.316 & 0.693 & 0.267 \\
\suitename{swahili-N\_of\_Poss\_D\_A\_V} & 0.624 & 0.006 & 2.068 & 0.270 & 0.742 \\
\suitename{swahili-N\_of\_Poss\_D\_AP\_ni\_AN} & 0.257 & 0.022 & 1.417 & 0.241 & 0.484 \\
\suitename{swahili-N\_of\_Poss\_D\_AP\_V\_ni\_AN} & 0.128 & 0.007 & 0.791 & -0.139 & 0.197 \\
\suitename{swahili-N\_of\_Poss\_D\_ni\_A} & 0.137 & 0.019 & 1.291 & -0.041 & 0.352 \\
\suitename{swahili-N\_of\_Poss\_D\_V} & 0.428 & 0.041 & 2.696 & 0.246 & 0.853 \\
\suitename{swahili-N\_of\_Poss\_V} & 0.351 & 0.039 & 1.949 & -0.703 & 0.409 \\
\end{tabular}
\caption{Slopes of best fit lines representing the relationship between accuracy (in percentage) and parameter count (in billions) for each model family on each test suite, given to three decimal places.}
\label{tab:4}
\end{table*}

\subsection{Robustness to intervening content}
\label{sec:robustness-to-intervening-content}

Prior studies have yielded different results on the stability of Transformer LMs to intervening constituents: \citet{wang-2021} found that increasing complexity of intervening material causes performance to degrade, \citet{hu-2020} found no significant performance degradation, and the BLiMP paper found that some Transformers are more prone to degradation than others.

\begin{figure}[H]
\centering
\includegraphics[width=0.95\linewidth]{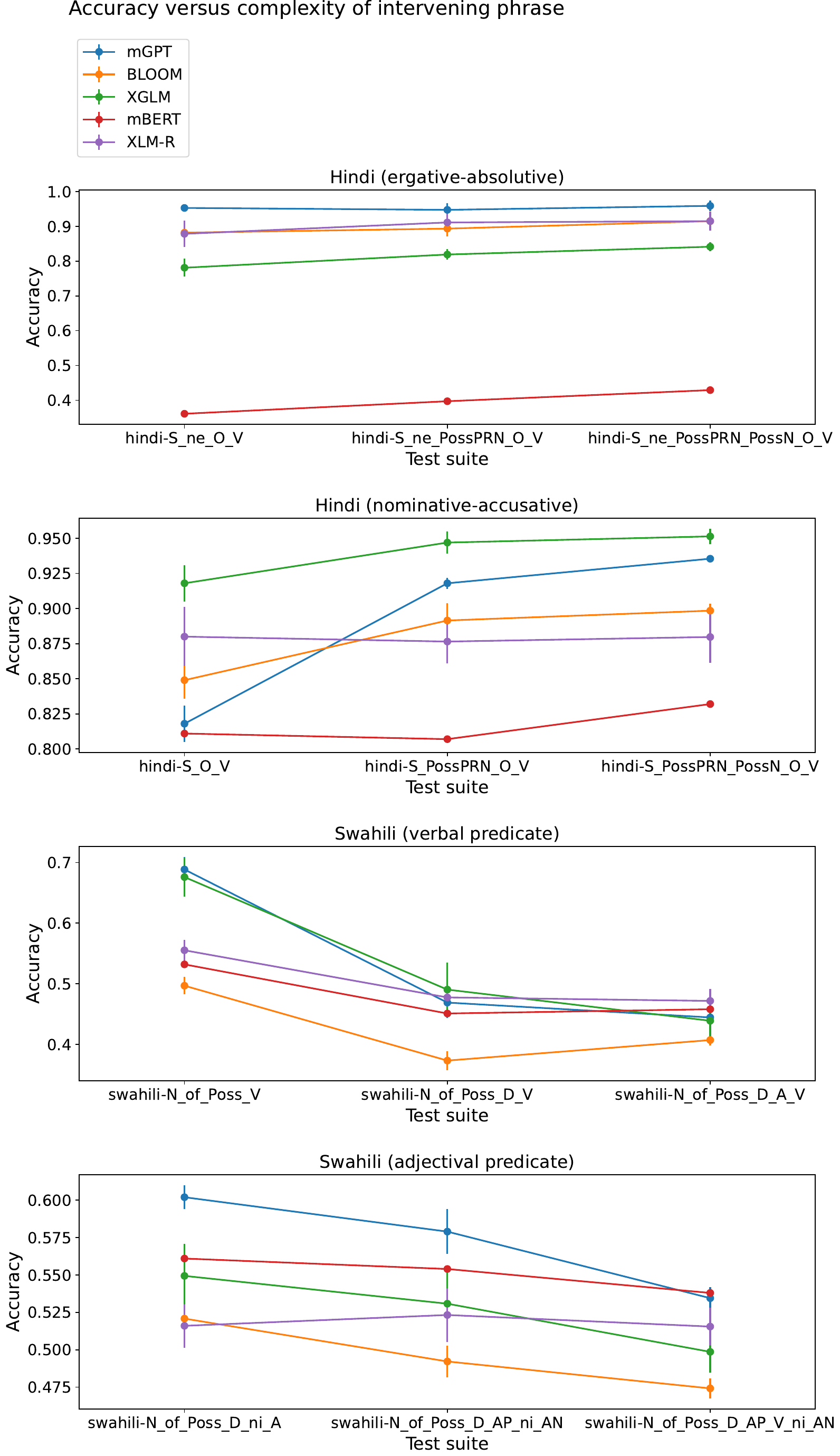}
\caption{Accuracy as a function of the complexity of the intervening constituent for Hindi and Swahili test suites. For models available in multiple versions, we show mean accuracy over versions; error bars denote 95\% confidence intervals on the standard error of the mean.}
\label{fig:intervening-content}
\end{figure}

Figure \ref{fig:intervening-content} shows accuracy as a function of the complexity of the intervening constituent for Hindi and Swahili test suites. It is visually apparent that the general trend is upward (i.e., no degradation at all) for Hindi but downward for Swahili. In Swahili, a particularly sharp drop in accuracy (minus 138.889 correct selections on average) results from the insertion of a demonstrative between the possessor and a verbal predicate. For both `preposition + noun + demonstrative' and `preposition + noun + demonstrative + adjective' PPs, accuracy is lower when the PP stands before a verbal predicate than before an adjectival predicate.

\section{Conclusion}
\label{sec:conclusion}

We assessed the ability of open-access multilingual Transformer LMs to form syntactic generalizations across three low-resource languages—Basque, Hindi, and Swahili. We found that models mostly performed well on Basque auxiliary agreement, albeit with challenges in sentences containing indirect objects, likely due to their relatively low frequency in training corpora. In Hindi, all LMs demonstrated a solid grasp of split ergativity except multilingual BERT, which failed to select the perfective aspect as the only grammatical aspect in sentences containing the ergative clitic. Noun class agreement in Swahili posed the greatest challenge, with models often performing near random guessing.

We hope that our work will motivate further investigations into LMs' linguistic knowledge in low-resource languages and will help LM developers identify and address areas for improvement, ultimately guiding the design of better LMs for low-resource languages and enabling fair access to high-quality NLP technologies for their speakers.

\section{Limitations}
\label{sec:limitations}

First, the present study focuses on one syntactic

\noindent phenomenon per language, which limits the generalizability of the findings. Future work could provide a more comprehensive analysis by considering multiple phenomena in each language and comparing performance on the same phenomenon across languages.

Second, since demographic data are self-reported on Prolific, we cannot be certain that the participants of our human validation study were true proficient speakers of the languages we investigated. Our method for verifying proficiency, the inclusion of two control minimal pairs in each dataset presented to human reviewers, sometimes allows non-proficient participants to pass: a participant who makes every selection by guessing selects the grammatical option in both control pairs with a 25\% probability.

Third, we were unable to find exact figures for the training dataset sizes of mGPT, XGLM (post-upsampling), and mBERT on Basque, Hindi, and Swahili, which limited our ability to analyze the relationship between performance and training dataset size.

\bibliography{custom}

\appendix

\section{Models Evaluated}
\label{sec:appendix}

\subsection{mGPT}
\label{sec:mgpt}

mGPT is an autoregressive model based on GPT-3 architecture introduced in \citet{sliazhko-2024}. It supports 61 languages and is trained on a combination of Wikipedia and Colossal Clean Crawled Corpus (C4). mGPT is available in two versions: mGPT\textsubscript{1.3B} (1,417,596,928 parameters) and mGPT\textsubscript{13B} (13,108,070,400 parameters).

\subsection{BLOOM}
\label{sec:bloom}

BLOOM is an autoregressive model developed over the course of a year-long open research workshop involving more than a thousand researchers \cite{bigscience-2022}. The model supports 46 natural languages and 13 programming languages and was trained on the ROOTS corpus, a composite collection of 498 Hugging Face datasets compiled by BigScience. BLOOM is available in six versions, ranging from 560 million to 176 billion parameters (see Table \ref{tab:bloom-versions}).

\begin{table}[h]
\centering
\begin{tabular}{c|c}
\textbf{Version} & \textbf{Number of parameters}\\
\hline
BLOOM\textsubscript{560M} & 559,214,592\\
BLOOM\textsubscript{1.1B} & 1,065,314,304\\
BLOOM\textsubscript{1.7B} & 1,722,408,960\\
BLOOM\textsubscript{3B} & 3,002,557,440\\
BLOOM\textsubscript{7.1B} & 7,069,016,064\\
BLOOM\textsubscript{176B} & 176,247,271,424\\
\end{tabular}
\caption{BLOOM versions.}
\label{tab:bloom-versions}
\end{table}

\subsection{XGLM}
\label{sec:xglm}

XGLM is an autoregressive model developed by Meta \cite{lin-2022}, trained on a corpus covering 68 Common Crawl snapshots. The model is available in five versions (see Table \ref{tab:xglm-versions}). XGLM\textsubscript{4.5B} supports 134 languages, while other versions support 30 languages.

\begin{table}[h]
\centering
\begin{tabular}{c|c}
\textbf{Version} & \textbf{Number of parameters}\\
\hline
XGLM\textsubscript{564M} & 564,463,616\\
XGLM\textsubscript{1.7B} & 1,732,907,008\\
XGLM\textsubscript{2.9B} & 2,941,505,536\\
XGLM\textsubscript{4.5B} & 4,552,511,488\\
XGLM\textsubscript{7.5B} & 7,492,771,840\\
\end{tabular}
\caption{XGLM versions.}
\label{tab:xglm-versions}
\end{table}

\subsection{Multilingual BERT}
\label{sec:mbert}

Multilingual BERT (mBERT) is the multilingual variant of BERT, an encoder-only Transformer developed by Google for the masked language modeling objective \cite{devlin-2019}. The model contains 177,974,523 parameters, was trained on Wikipedia, and supports 104 languages with the largest Wikipedias at the time of training.

\subsection{XLM-RoBERTa}
\label{sec:xlmr}

XLM-RoBERTa (XLM-R) is a masked model developed by Facebook with the goal of improving upon previous state-of-the-art models such as mBERT \cite{conneau-2020}. The model was trained on a cleaned version of the Common Crawl corpus that encompasses 100 languages, including 93 natural languages, the constructed language Esperanto, five romanizations of South Asian languages typically written in non-Latin scripts, and a variant of Burmese written in the non-Unicode-compliant Zawgyi font. XLM-RoBERTa is available in four versions, as outlined in Table \ref{tab:xlmr-versions}.

\begin{table}[H]
\centering
\begin{tabular}{c|c}
\textbf{Version} & \textbf{Number of parameters}\\
\hline
XLM-R\textsubscript{Base} & 278,295,186\\
XLM-R\textsubscript{Large} & 560,142,482\\
XLM-R\textsubscript{XL} & 3,482,741,760\\
XLM-R\textsubscript{XXL} & 10,712,994,816\\
\end{tabular}
\caption{XLM-RoBERTa versions.}
\label{tab:xlmr-versions}
\end{table}

\end{document}